# Deep Learning for Semantic Segmentation of 3D Ultrasound Data

Chenyu Liu[1,2][0009-0003-5242-6150], Marco Cecotti[1][0000-0002-4457-3631], Harikrishnan Vijayakumar[1][0000-0001-7718-6614], Patrick Robinson[2][0009-0008-5632-0595], James Barson[2][0009-0006-6732-6437], and Mihai Caleap[2][0000-0001-8960-7762]

[1] Cranfield University, College Road, Wharley End, Bedford MK43 0AL, U.K.
[2] Calyo Limited, Future Space, Filton Road, Bristol BS34 8RB, U.K.
`mihai@calyo.com`

**Abstract.** Developing cost-efficient and reliable perception systems remains a central challenge for automated vehicles. LiDAR and camera-based systems dominate, yet they present trade-offs in cost, robustness and performance under adverse conditions. This work introduces a novel framework for learning-based 3D semantic segmentation using Calyo Pulse, a modular, solid-state 3D ultrasound sensor system for use in harsh and cluttered environments. A 3D U-Net architecture is introduced and trained on the spatial ultrasound data for volumetric segmentation. Results demonstrate robust segmentation performance from Calyo Pulse sensors, with potential for further improvement through larger datasets, refined ground truth, and weighted loss functions. Importantly, this study highlights 3D ultrasound sensing as a promising complementary modality for reliable autonomy.

## 1    Introduction

Robust 3D perception is essential for autonomous navigation, requiring sensors that deliver dense environmental data under diverse conditions [1, 2]. While LiDAR and radar are widely used, their high costs and limitations in adverse weather motivate the search for alternative solutions. In-air 3D ultrasound sensors such as Calyo Pulse [3] offer a promising low-cost option with strong robustness to environmental factors, though their range is constrained by acoustic travel time (typically limited to 15-25m depending on the speed of the vehicle) (Fig. **1**a). This research, part of the *Driven by Sound* project [4], explores ultrasound-based perception for autonomous driving, with an emphasis on 3D semantic segmentation using LiDAR-informed deep learning.

For experimentation, two Calyo Pulse sensors were mounted on a Kia Niro EV (Fig. **1**b-d). To support autonomous navigation, we propose a learning-based 3D semantic segmentation approach that improves environmental understanding by training a 3D U-Net on ground truth derived from LiDAR data. This approach assigns a class label to every voxel in the detection volume, generating a dense semantic map. Compared with bounding boxes, semantic maps provide more precise object shape representation, although they do not distinguish between individual objects of the same class, which limits tracking capability.

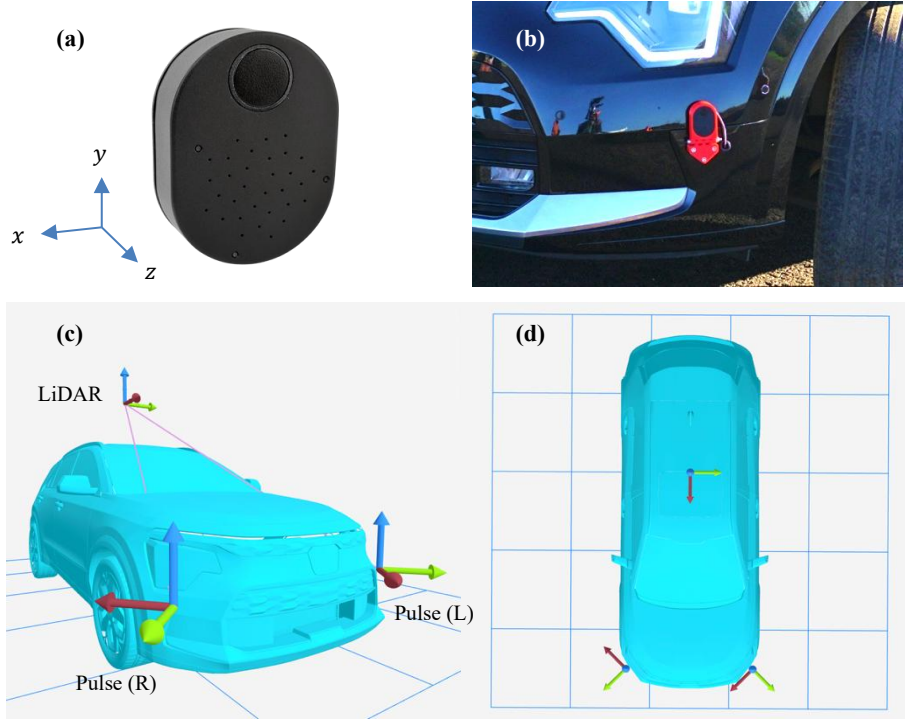

**Fig. 1.** Calyo Pulse 3D ultrasound sensor (a); Pulse fitted on the test vehicle (b); Installation position of LiDAR and Calyo Pulse sensors (c-d).

The application of 3D semantic segmentation to ultrasound data represents a novel and unexplored direction for autonomous vehicles. Prior work in this area is limited, and early-stage experiments by the authors highlighted challenges such as cluttered detection images that obscure object boundaries. The proposed framework demonstrates that these limitations can be mitigated through advanced pre-processing and deep learning.

Importantly, this work shows that compact 3D ultrasound sensors such as Calyo Pulse can serve as a viable complement to traditional LiDAR, radar, and camera systems. By providing resilient 3D spatial information under conditions where optical and radio-frequency sensors often fail, ultrasound sensing opens a new pathway for redundancy and safety in autonomous navigation (see e.g. [5]). This study therefore contributes not only an algorithmic approach to ultrasound-based semantic segmentation, but also evidence that a new class of 3D perception sensors can enhance robustness in real-world autonomy.

## 2 Related Work

### 2.1 Sensor Hardware: Calyo Pulse

Calyo Pulse is a compact in-air 3-D ultrasound sensor designed for volumetric perception. The modular system employs a transceiver element (Tx), nominally centered at 40 kHz but configurable for other frequencies, and a broadband receiver (Rx) array with a wide spectral response (100 Hz - 80 kHz), see Fig. **2**. This configuration enables flexible operation across different acoustic regimes and supports high-fidelity time-of-flight and amplitude measurements. The hardware is fully solid-state, with no moving parts, and engineered for ruggedised operation in automotive and industrial environments. The enclosure is manufactured from anodised aluminium alloy, providing mechanical stability, vibration resistance, and environmental protection. Power and data are provided through a USB-C interface (5 V, 0.15 A) with total power consumption under 1 W. The device outputs raw Rx waveforms, 3D point clouds, and 2-D intensity projections accessible *via* the Calyo Sensus SDK, which supports real-time beamforming, data capture, and visualisation [3].

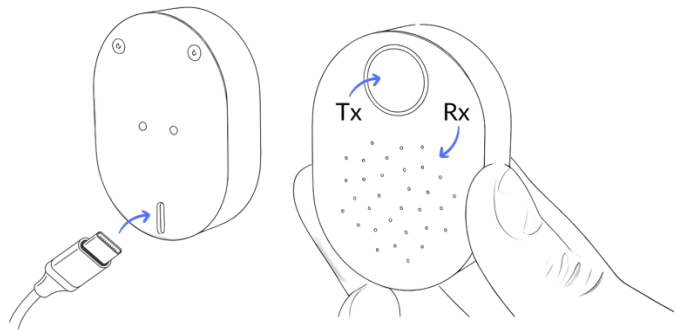

**Fig. 2.** Calyo Pulse 3D ultrasound sensor featuring a transceiver Tx and a configurable microphone array Rx, and embeds a highspeed microcontroller for real-time data conditioning and acquisition.

### 2.2 3D Beamforming

Beamforming enhances signal coherence and reduces noise by combining signals from multiple array elements. Traditional delay-and-sum (DAS) methods [6] are straightforward but limited in resolution. Adaptive approaches such as minimum variance distortionless response (MVDR) [7] improve resolution and noise suppression by minimizing output power while preserving distortionless target response. Fig. **3** compares DAS and MVDR outputs from a Calyo Pulse sensor in a representative environment using the Calyo Sensus SDK. Each Calyo Pulse sensor incorporates a 32-channel receive (Rx) array, and the SDK provides both DAS and MVDR implementations. For a complete overview of the Pulse hardware and Sensus software suite, see [3].

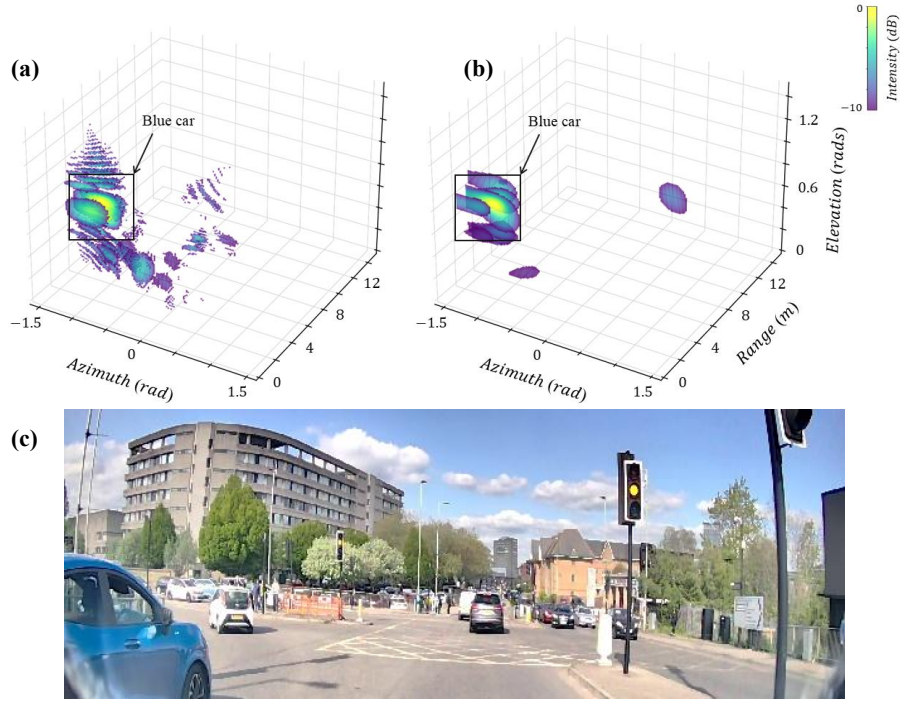

**Fig. 3.** Comparison of the point spread function between DAS (a) and MVDR (b) using the left-mounted Pulse sensor, and corresponding camera frame (c).

### 2.3   3D Semantic Segmentation

Image segmentation enhances environmental understanding by partitioning data into semantically labelled regions. Early methods relied on sliding-window classification [8], later replaced by fully convolutional networks (FCNs) [9], which allow dense, pixel-level predictions. Architectures such as U-Net [10] have become foundational, while Dice loss [11] is widely used to mitigate severe class imbalance. Although most prior work has focused on biomedical or optical imagery [12], these methods can be adapted to volumetric in-air ultrasound data.

### 2.4   Dataset and Ground Truth Annotation

LiDAR remains the standard for generating ground truth *via* rule-based segmentation [13, 14], often refined with human-in-the-loop adjustments [15]. This study integrates LiDAR-based segmentation with manual refinement to build training datasets. The dataset was collected on urban roads in Bedford, U.K., during peak hours, including slow-moving traffic, roundabouts, and pedestrian crossings, conditions that provided a representative testbed for real-world perception challenges.

## 3      Methodology

### 3.1      Hardware Platform

Experiments used a 2022 Kia Niro EV fitted with two Calyo Pulse sensors mounted on the front bumper, one LiDAR, multiple cameras, and an inertial navigation system (INS). Data collection focused on 3D ultrasound and LiDAR inputs, with installation positions shown in Fig. **1**c-d.

### 3.2      3D Ultrasound Intensity Map Generation

Following beamforming, raw RF signals were processed to generate 3D intensity maps. The maps were first represented in spherical polar coordinates $(r, \theta, \varphi)$ and then converted into Cartesian space $(x, y, z)$ for interpretability. The full setup is displayed in Table **1**. The processing pipeline also included Constant False Alarm Rate (CFAR) filtering to suppress noise, and linear range compensation to enhance visibility of distant objects. Example outputs are shown in Fig. **4** where clutter is significantly reduced after CFAR application.

**Table 1.** Settings for ultrasound sensor output.

| | |
|---|---|
| **Beamforming** | MVDR |
| **Range $r$** | $0 - 12\ m$, step $0.125\ m$ |
| **Azimuth $\theta$** | $[-90°, 90°]$, 64 bins |
| **Elevation $\varphi$** | $[-10°, 30°]$, 64 bins |
| **Processing** | CFAR, ground removal, ego vehicle removal, clustering |

The final processed 3D intensity maps were stored in structure format with dimensions of $64 \times 96 \times 64$ (height, depth, width), generated *via* the Calyo Python SDK [16].

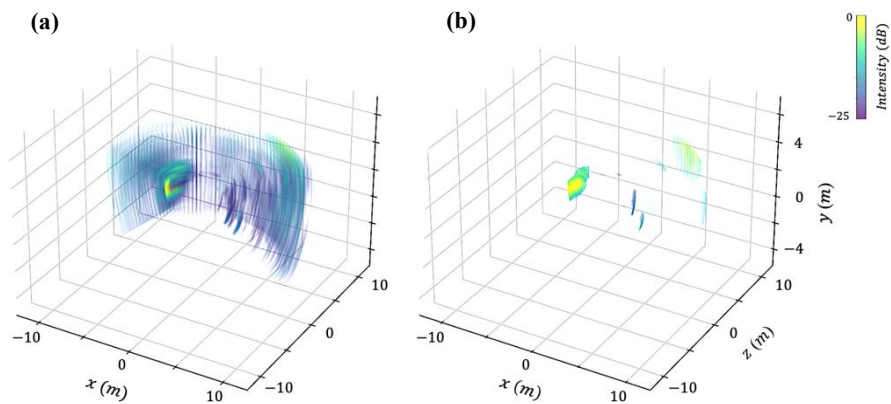

**Fig. 4.** 3D volumetric intensity map before (a) and after (b) processing.

### 3.3   LiDAR Ground Truth Segmentation Masks

**Semi-Automatic Annotation.** Ground truth (GT) annotations were generated using a segmentation-based bounding box generation algorithm combined with rule-based cleaning and filtering, adapted from [17]. Key parameters are summarized in Table **2**. The initial bounding boxes were further refined manually to ensure accuracy, particularly in cases of partial occlusion or irregular object shapes. Representative results are shown in Fig. **5**.

**Table 2.** Ground truth annotation setup.

| Process Step | Parameter | Value / Threshold |
| --- | --- | --- |
| ROI Settings | Forward range ($z$) | 0 $m$ to 12 $m$ |
| Max Epochs | Horizontal range ($x$) | $-12\ m$ to 12 $m$ |
| Mini Batch Size | Vertical range ($y$) | $-2\ m$ to 6 $m$ |
| Ground Removal | Removal distance | 0.3 $m$ |
| Ego Vehicle Removal | Removal radius | 2 $m$ |
| Clustering | Interval threshold | 0.7 $m$ |
| Rule-Based Filtering | Length (L) | 0.2 $m$ to 8 $m$ |
| Learning Rate Drop Factor | Width (W) | 0.5 $m$ to 3 $m$ |
| | Height (H) | 0.5 $m$ to 3 $m$ |
| | L/W ratio | 0.5 to 5 |
| | L/H ratio | $\geq 0.4$ |
| Final Output | Frame count | 1146 |
| | Bounding Box Attributes | $(cx, cy, cz, l, w, h, \theta)$ |

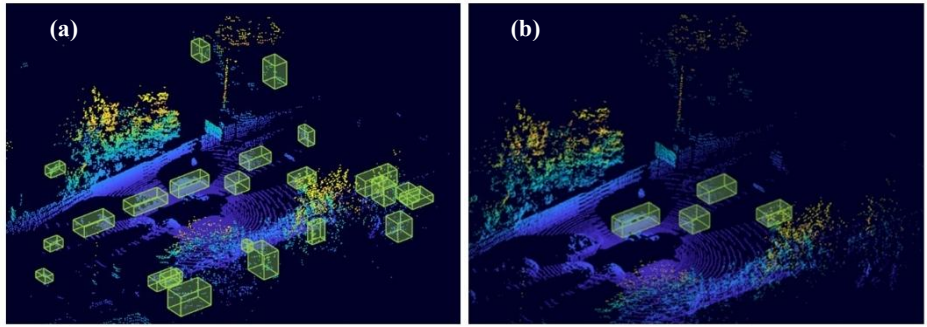

**Fig. 5.** Bounding boxes before (a) and after (b) refinement.

**Occupancy Ground Truth Masks.** To facilitate semantic segmentation, bounding boxes were projected onto an occupancy grid. The dataset produced 821 aligned frames of LiDAR-derived ground truth with corresponding ultrasound intensity maps. Projection accuracy was ensured using extrinsic calibration parameters, shown in Table **3**, which aligned the ultrasound coordinate system with the LiDAR frame. An example of the aligned masks is shown in Fig. **6**.

**Table 3.** Extrinsic calibration parameters (relative to LiDAR).

| Frame ID (child) | $x\,(m)$ | $y\,(m)$ | $z\,(m)$ | $\theta\,(°)$ | $\varphi\,(°)$ |
|---|---|---|---|---|---|
| Pulse (L) | $-0.8$ | $-1.5$ | 2.5 | $-135$ | 5 |
| Pulse (R) | 0.8 | $-1.5$ | 2.5 | $-45$ | 5 |

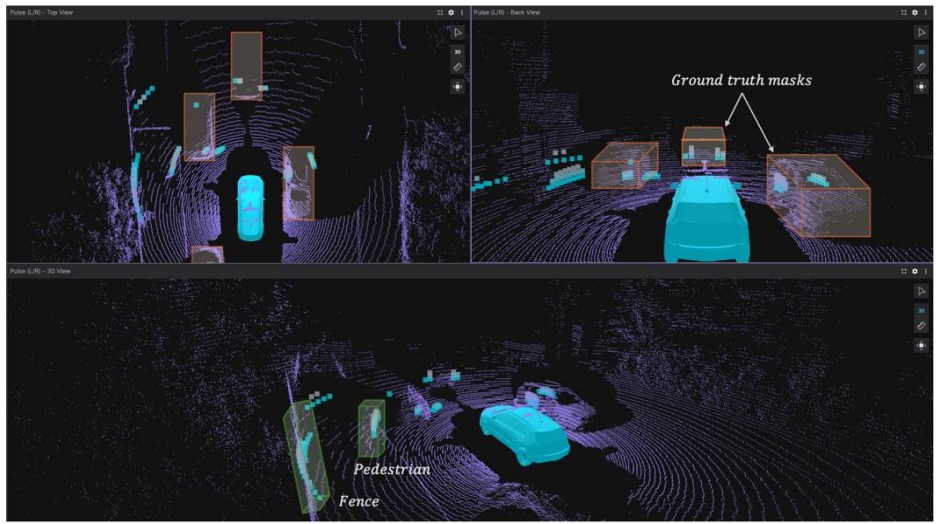

**Fig. 6.** Top: BEV and 3D view for ultrasound intensity maps and LiDAR ground truth for both Pulse sensors combined. Bottom: An example of valid ultrasound returns where objects are not labelled.

### 3.4    3D U-Net Architecture and Training

**Pre-Processing.** The left and right ultrasound datasets were combined, producing 1586 frames of input data and corresponding GT masks. These were split $80/20$ into training (1268 frames) and validation (318 frames). Empty frames with no GT labels were removed to improve training efficiency. The loss function was defined as $1 - mean$(Dice), rewarding the overlap between prediction and ground truth.

**Network Architecture.** A 3D U-Net was implemented, with two encoder and two decoder stages, reducing depth relative to the original four-stage architecture [10] discussed in Section 2.2. The input intensity map ($64 \times 96 \times 64 \times 1$) was progressively downsampled, with feature channels doubled at each stage. The bottleneck layer expanded to 64 channels before upsampling through symmetric decoder stages. Shortcut connections preserved spatial detail, and the output layer generated voxel-wise predictions across two classes ("background" and "object"), normalized using Soft-Max. This process is illustrated in Fig. **7**.

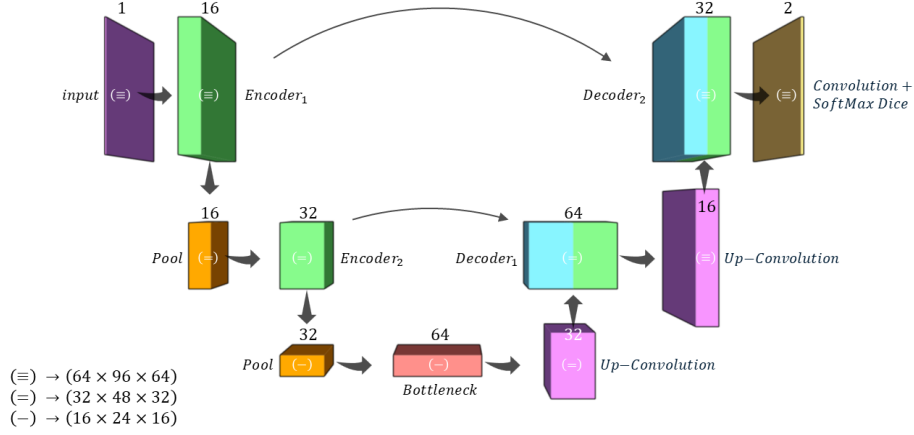

**Fig. 7.** Main stages of the 3D U-Net used in this study. The number of channels is denoted above each voxel tile/feature map.

**Training Setup.** Training was performed using the Adam optimizer with a piecewise learning rate schedule (Table **4**). Networks were validated once per epoch with results recorded for quantitative analysis.

## 4        Results and Discussion

This section presents the quantitative and qualitative analysis of the proposed model. Unless otherwise noted, results are based on combined volumetric intensity maps from the left and right Pulse sensors, processed with CFAR and range compensation. Empty GT frames were removed prior to training to ensure consistency.

**Table 4.** Training setup for the 3D U-Net.

| Option | Value |
|---|---|
| Optimizer | Adam |
| Max Epochs | 20 |
| Mini-Batch Size | 1 |
| Initial Learning Rate | 0.0001 |
| Shuffle | Every epoch |
| Learning Rate Schedule | Piecewise |
| Learning Rate Drop Period | 10 epochs |
| Learning Rate Drop Factor | 0.3 |

### 4.1    Quantitative Analysis

Fig. **8** shows training and validation curves. Accuracy rose sharply to 98% within three epochs, stabilizing thereafter without overfitting. Training loss dropped from 0.97 to 0.44, while validation loss plateaued at ~0.54 indicating limited generalization.

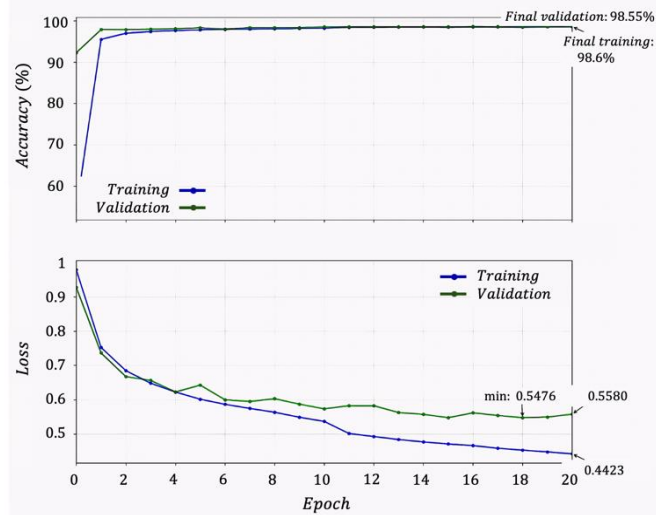

**Fig. 8.** Training and validation progress over 20 epochs.

Next, the trained model was tested on the dataset from the left Pulse sensor with 821 frames to assess the generalization to unprocessed datasets. Per-class evaluation (Table **5**) reveals strong background detection (> 99%) and reduced clutter, but moderate recall (62.69%) for objects. Precision (72.64%) suggests a conservative model that favors clean predictions at the expense of some missed detections.

**Table 5.** Per-class evaluation of the baseline model.

| Class | Total Voxels | Recall | Precision | F1 (Dice) | IoU |
|---|---|---|---|---|---|
| Background | 317,819,270 | 99.63% | 99.41% | 99.52% | 99.05% |
| Object | 5,011,066 | 62.69% | 72.64% | 67.30% | 50.72% |

### 4.2    Qualitative Analysis

Qualitative results highlight both strengths and limitations. Front camera images, processed using YOLOv8 [18], were used to cross-check detections and highlight cases where ultrasound-based predictions exceeded the quality of ground truth annotations. As shown in Fig. **9**, the model successfully detected vehicles directly ahead (true positives), while missing occluded objects (false negatives). Fig. **10** demonstrates a case where the model predicted an object not labelled in the GT, corroborated by camera evidence, indicating that predictions may sometimes exceed the quality of GT annotations.

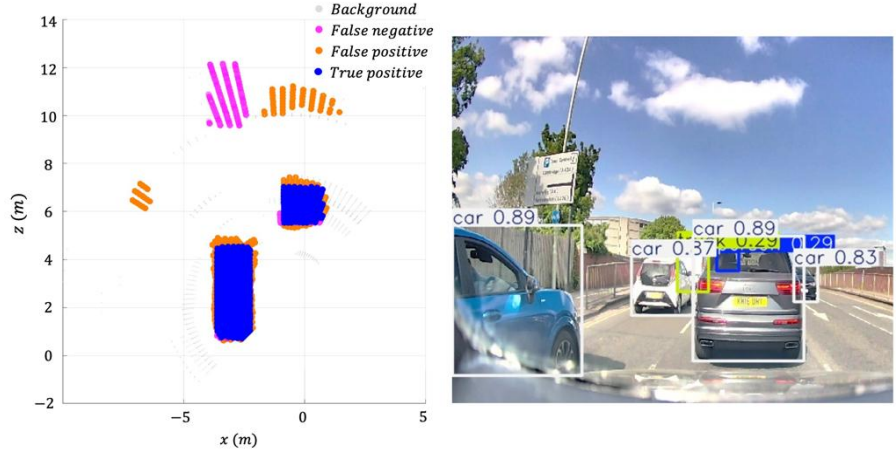

**Fig. 9.** True positive and false negative case (33$^{rd}$ frame).

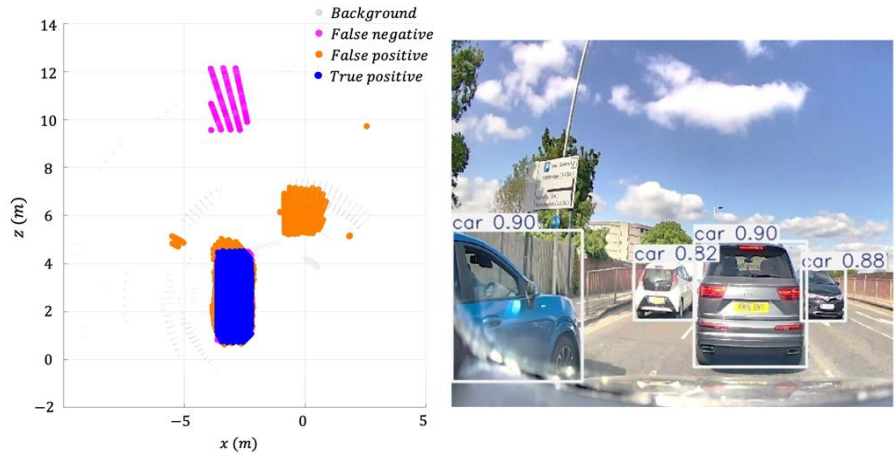

**Fig. 10.** True positive, false negative, and false positive case (34$^{th}$ frame).

**Discussion**. A limitation of the present study lies in the scope of the ground-truth annotation. The LiDAR dataset was labelled exclusively with bounding boxes around vehicles, which served as the sole object class for segmentation. However, the ultrasound data naturally contains reflections from other environmental features such as pedestrians, curbs, poles, fences, and vegetation. These additional scattering sources appear in the ultrasound intensity maps but are absent from the LiDAR-derived ground truth. Consequently, the U-Net may treat valid ultrasound returns from non-vehicle objects as background noise (see Fig. **6**), leading to apparent false positives or incomplete segmentation. This mismatch between sensor perception and annotation scope suggests that the true generalization capability of the network may be underestimated and highlights the need for broader multi-class ground-truth labelling to capture the full acoustic complexity of real-world scenes.

### 4.3    Ablation Studies

This section evaluates how specific design choices influence the performance of the proposed model. Building on the baseline described above, two factors are examined independently: (i) the impact of filtering empty frames, and (ii) the contribution of CFAR pre-processing.

**Filtering.** To evaluate the role of filtering, models trained with and without empty-frame removal were compared (Table **6**). Filtering improved recall, precision, and F1 by ∼5%, confirming its effectiveness.

**Table 6.** Impact of filtering on performance.

| Model | Recall | Precision | F1 (Dice) | IoU | Best Val. Loss |
|---|---|---|---|---|---|
| Filtered | 62.69% | 72.64% | 67.30% | 50.72% | 0.52 |
| Unfiltered | 57.66% | 67.83% | 62.33% | 45.28% | 0.57 |

**CFAR.** The effect of CFAR pre-processing is shown in Table **7**. While disabling CFAR slightly reduced validation loss, it decreased both recall and precision, suggesting mild overfitting and poorer generalization.

**Table 7.** Impact of CFAR on segmentation performance.

| Model | Recall | Precision | F1 (Dice) | IoU | Best Val. Loss |
|---|---|---|---|---|---|
| CFAR-enabled | 62.69% | 72.64% | 67.30% | 50.72% | 0.52 |
| CFAR-disabled | 58.37% | 71.36% | 64.22% | 47.29% | 0.50 |

## 5    Conclusions

This study introduced one of the first demonstrations of 3D semantic segmentation using in-air ultrasound sensors (Calyo Pulse) guided by LiDAR-informed deep learning. The proposed framework combined beamforming of raw RF signals, CFAR filtering, and range compensation, followed by LiDAR-assisted ground truth generation and training of a 3D U-Net.

Experimental results showed promising segmentation capability: recall of 62.69% and F1 of 67.30%, with strong background detection (> 99%) and effective clutter reduction. Ablation studies confirmed that filtering and CFAR processing significantly improved performance. While challenges remain, particularly dataset size, class imbalance, and refinement of ground truth, this work demonstrates the feasibility of ultrasound-based volumetric segmentation for autonomous perception. This work also serves as a baseline model for future improvements.

The contribution is twofold: methodologically, it provides a reproducible framework for processing ultrasound volumetric data (point clouds) with deep learning; conceptually, it highlights 3D ultrasound as a complementary perception modality. Unlike LiDAR, radar, and cameras, which can fail in fog, dust, rain, or poor lighting, ultrasound offers resilient, low-power sensing.

Looking ahead, integrating 3D ultrasound with existing sensing modalities offers a path to robust multimodal autonomy. With larger datasets and refined annotation, this work can extend from binary classification to multi-class detection and tracking. Beyond road vehicles, potential applications include mobile robots, rail safety, and industrial automation. More broadly, compact 3D ultrasound devices may underpin the next generation of foundational perception technologies in robotics.

**Acknowledgments.** This research was carried out as part of the *Driven by Sound* project, led by Calyo and supported by UK Research and Innovation under grant number 10059986. The authors gratefully acknowledge the contributions of project partners and collaborators who provided technical insights and support during development and testing.

**Disclosure of Interests.** Calyo Limited, which employs several of the authors, is the developer and manufacturer of the Calyo Pulse devices used in this study.

**Open Access Policy.** For the purpose of open access, the authors have applied a Creative Commons Attribution (CC BY) license to any Accepted Manuscript version arising.

## Appendix

The hardware specifications of the workstation used for data processing and CNN training are listed in Table **8**.

**Table 8.** Computer hardware specifications for experimental setup.

| Component | Specification |
|---|---|
| CPU | Intel Core i7-14650HX (16-Core) |
| GPU | NVIDIA GeForce RTX 4060 Laptop GPU (8 GB VRAM) |
| System Memory (RAM) | 32 GB Hynix DDR5 @ 5600MHz (2 x 16 GB) |
| Storage | 1 TB UMIS NVMe SSD |

The project was implemented and executed on an Ubuntu 22.04.5 (x86_64) platform. The accompanying software stack is summarized below:

- System & Drivers:
  - NVIDIA Driver: 570.133.07
  - CUDA Version: 12.8
- Calyo SENSUS Software Suite (v4.2.3):
  - Prerequisite: SENSUS 3rd-party package (v2.1.0)
  - Tooling: SENSUS Viewer (v3.4.4)
  - Core SDKs:
    - Calyo Python SDK (v1.1.0, CUDA-enabled).

Calyo ROS SDK (v2.4.4, CUDA-enabled). Note: A specific pre-release version was utilised to ensure correct frame extraction from ROS bags.